# Generation of SAR Image for Real-life Objects using General Purpose EM Simulators

Amit Kumar Mishra[1] and Bernard Mulgrew[2]

[1]ECE Department, Indian Institute of Technology Guwahati, India, [2]IDCOM, University of Edinburgh, UK.

## Abstract

In the applications related to airborne radars, simulation has always played an important role. This is mainly because of the two fold reason of the unavailability of desired data and the difficulty associated with the collection of data under controlled environment. A simple example will be regarding the collection of pure multipolar radar data. Even after phenomenal development in the field of radar hardware design and signal processing, till now the collection of pure multipolar data is a challenge for the radar system designers. Till very recently, the power of computer simulation of radar signal return was available to a very selected few. This was because of the heavy cost associated with some of the main line electro magnetic (EM) simulators for radar signal simulation, and secondly because many such EM simulators are for restricted marketting. However, because of the fast progress made in the field of EM simulation, many of the current generic EM simulators can be used to simulate radar returns from realistic targets. The current article expounds the steps towards generating a synthetic aperture radar (SAR) image database of ground targets, using a generic EM simulator. It also demonstrates by the help of some example images, the quality of the SAR image generated using a general purpose EM simulator.

## 1. Introduction

There have been some reports in the open literature [1-9], on the use of specific electro magnetic (EM) simulation computational tools, for generating synthetic aperture radar (SAR) images, of modeled targets. The major need for such an approach [10,11] is to generate a database of radar signatures, when field-trials are not economical or impractical. Lack of data and field trials, is a major bottleneck for research into futuristic radar systems. For example multistatic radar systems, and fully polarimetric radar systems. Limited theoretical analysis has been the traditional solution in such cases. However, with the increase in the computing power available to researchers and newer methods of simulation, computer aided electro-magnetic (EM) simulation has been proving itself as another alternative, in the recent years. Although computer aided simulations can not replace theoretical analysis, it can certainly complement a purely mathematical analysis. In certain usages, like the study of multistatic radar systems, fully polarimetric radar systems and noise radars, computer aided simulations have the potential to supplement pure-theoretical analysis. There are certain computer aided EM simulators, which have been reported to be efficient at the task of radar signal simulation in general and SAR image simulation in particular [12-15]. Most such tools are restricted and are not available for a wider research community. However, the current generation generic EM simulation tools (By generic EM simulation tool, we mean the range of commercial tools which are mostly available in free market and whose main advertised use is regarding antenna design and RF communication system simulation.) [16,17] are coming with enough power so that many of them can be used in radar system simulation. One such new EM simulation FEKO [17] has been used in the experiments reported in this article. Recently, bistatic radar systems have drawn considerable interest and research efforts [1,18-22]. The integration of bistatic systems in airborne and satellite systems, is a field of active research. The study of the intricacies of bistatic imaging and the usages of bistatic images, need bistatic synthetic aperture radar (SAR) images collected in a controlled environment. In the current article, the generation of bistatic SAR dataset of ground targets, was taken as the problem to be tackled using the general purpose EM simulator FEKO. To verify the results, a detailed analysis of a sample SAR image, is also presented. Even though some works in the open literature have come up which have used simulated radar signals, mostly they have used the restricted EM simulation tools. To the author's knowledge, except for the one paper by the author [23], there has come no work in the open literature, describing the use of a general purpose EM simulator for such purposes. The next section describes some of the basic experiments done on the generic EM simulator FEKO, to test its suitability for the present application. This is followed by a description of the ground target CAD model generation steps, and then a section describing the SAR image formation steps. The next section analysis a sample SAR image. The article ends with a summary of the work.

## 2. Experiments Done on the EM Simulator for Analyzing Forward Scattering and Shadow Effects

In this section, a few key experiments will be described, which gave the confidence that FEKO could be used for the purpose of simulating radar images. These experiments are in addition to the range of other simulations reported in the FEKO users manual. In simulating the signatures of ground targets, the ground itself plays an important role. Hence, the capabilities of an EM simulator in dealing with ground reflection, forward scattering and in generating a shadow region on the ground, are of importance. In this section the forward scattering and shadow generation capabilities of FEKO will be demonstrated. The ground reflection capabilities will be more evident in section V, where a typical SAR image will be analyzed in detail. In an experiment to examine the forward scattering capability, the CAD model of a long rectangular perfectly electrical conducting (PEC) prism was modeled [Figure 1]. The size of the rectangular cylinder was 1 m × 1 m × 10 m. The frequency of the wave transmitted, was fixed at 1 GHz. The transmitter was fixed at 0° elevation and 45° azimuth, and the bistatic radar cross section (RCS) was measured round the target at 0° elevation Figure 2. The model and the bistatic RCS observed are shown in Figure 1 and Figure 3.

As can be seen from Figure 3, there are three peaks in the RCS plot, the first and the third corresponding to the specular-reflection, and the second peak corresponding to the forward scattering returns Figure 1. This demonstrates FEKO's ability in simulating forward scattering. In testing the shadow region simulation abilities, a simple experimental set-up was designed. The CAD model is shown in Figure 4. In this setup, a wall like structure was modeled on the ground. The transmitter was fixed at azimuth 0° and elevation 0° so as to get a far field approximation at the structure. The surface current intensity on the ground was mapped Figure 4.

It can be observed that the surface current distribution shows a shadow region behind (w.r.t. the transmitter) the structure.

Fringes of surface current, due to interference, can also be seen on the current-map.

These experiments showed FEKO's ability to simulate forward scattering and shadow regions.

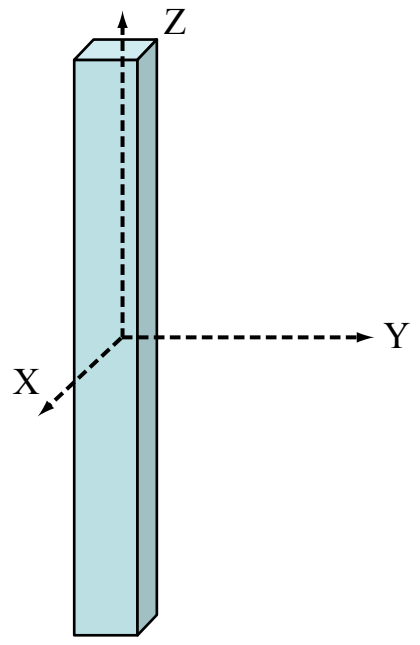

**Figure 1:** Model of the PEC rectangular prism.

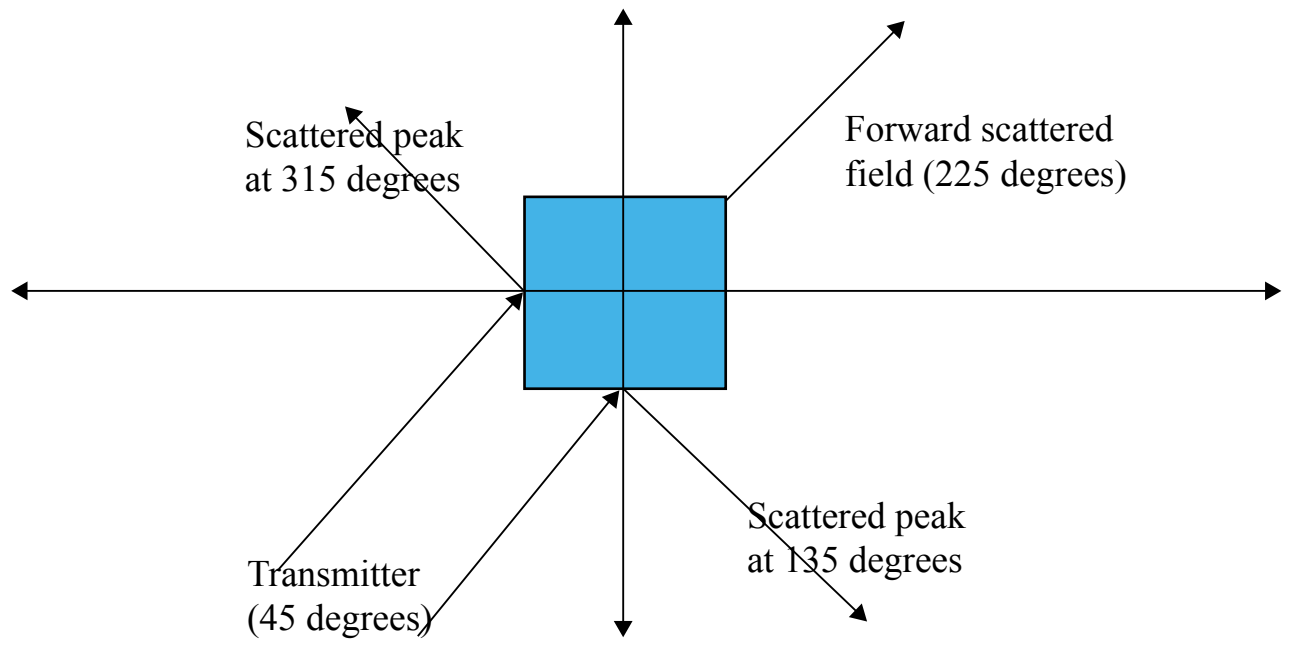

**Figure 2:** Cross section of the rectangular PEC prism, and explaining the peaks found in the bistatic RCS plot.

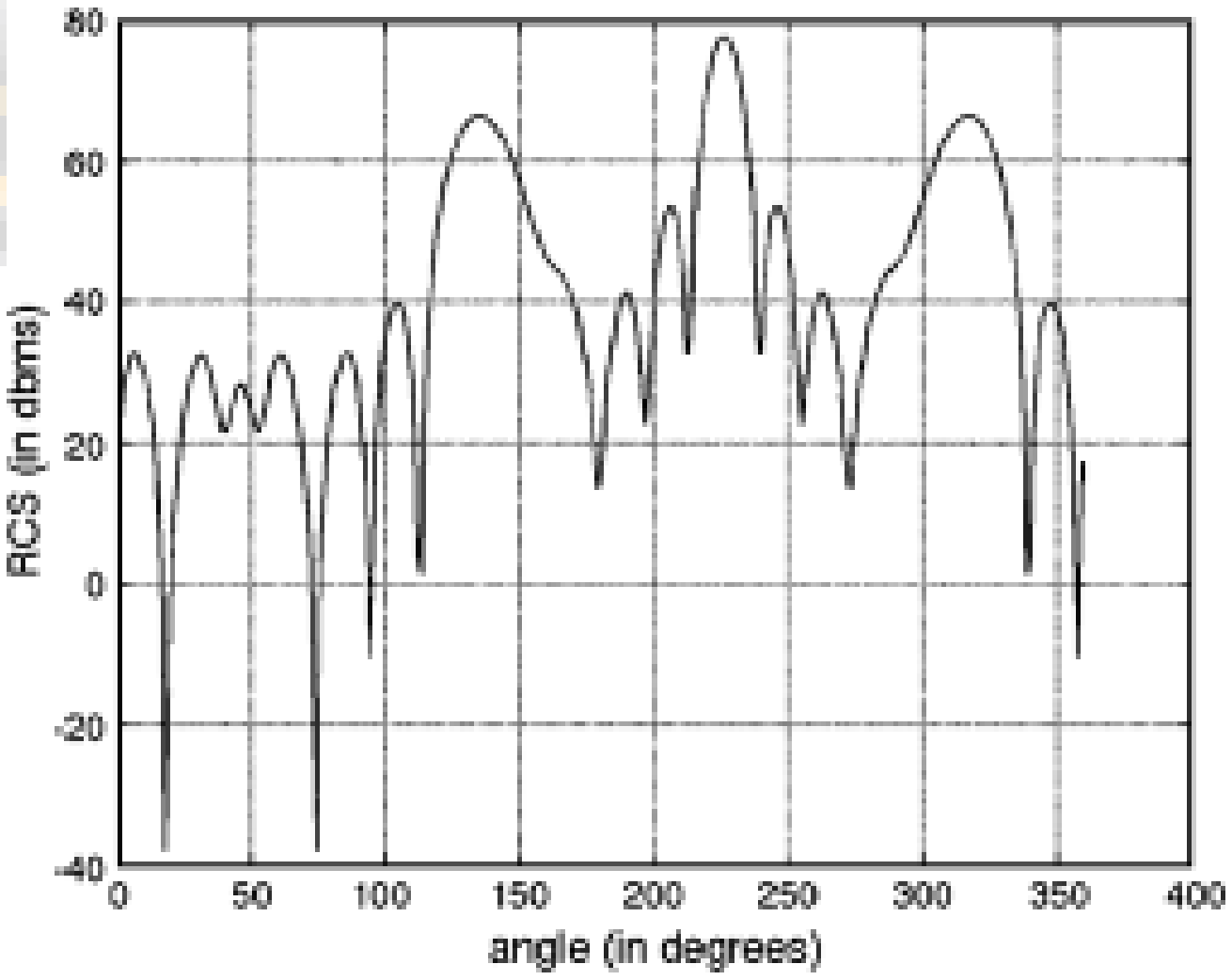

**Figure 3:** Bistatic RCS (in dbsm) versus azimuth angle in degrees.

## 3. CAD Modeling for EM Simulation

Military ground targets have highly involved and complex surfaces and features. Hence, to model even a fairly approximate ground target, it needs considerable expertise from the modeller and a highly efficient CAD tool. A major problem here is the issue of surface alignment. Like many of the other EM simulation tools, FEKO replaces the surfaces with small triangular facets. The dimension of these facets should be an order less than the central frequency wavelength. For two mutually touching triangular facet surfaces, the respective sides of the triangles should exactly match each other. This strict condition is often extremely difficult to satisfy for complex structures. In the current work instead of modeling the finer details of a target, the more prominent features of the target are modeled.

Such features are termed as the *classifiable features*[1] For example, the turret and the canon of a main battle tank,

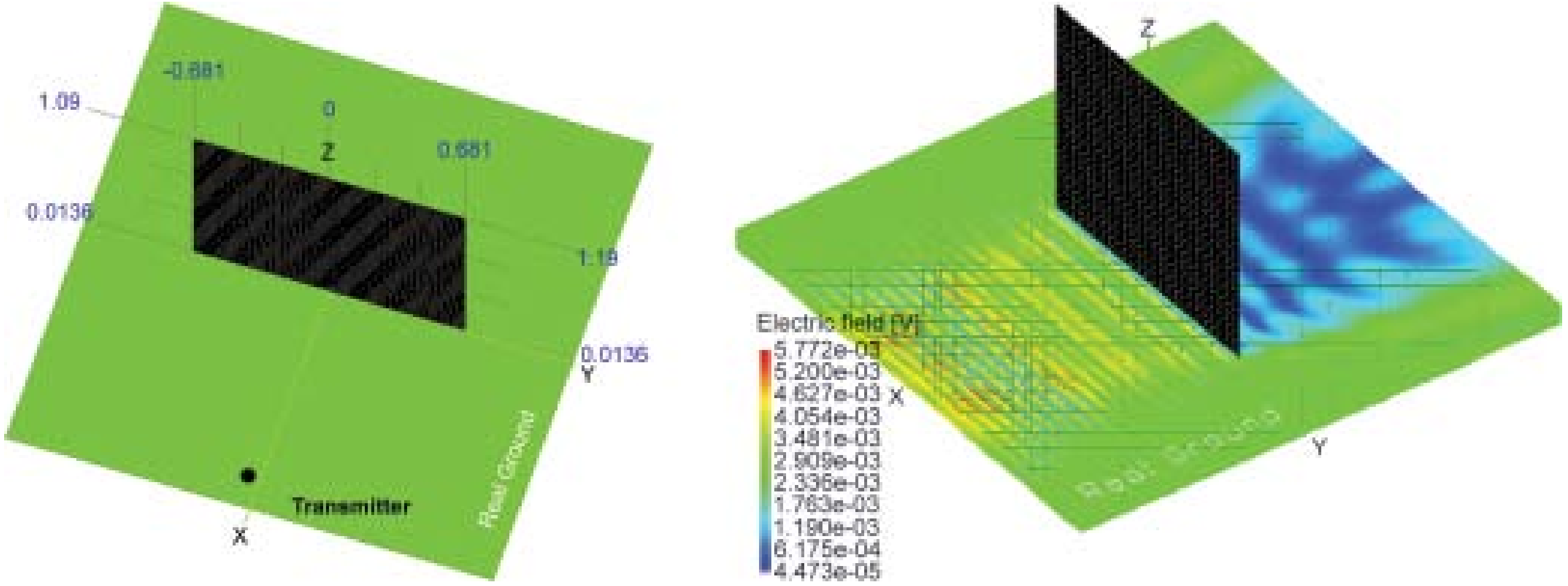

**Figure 4:** FEKO model of the setup and the surface current on the ground (red represents highest density of current and violet the least).

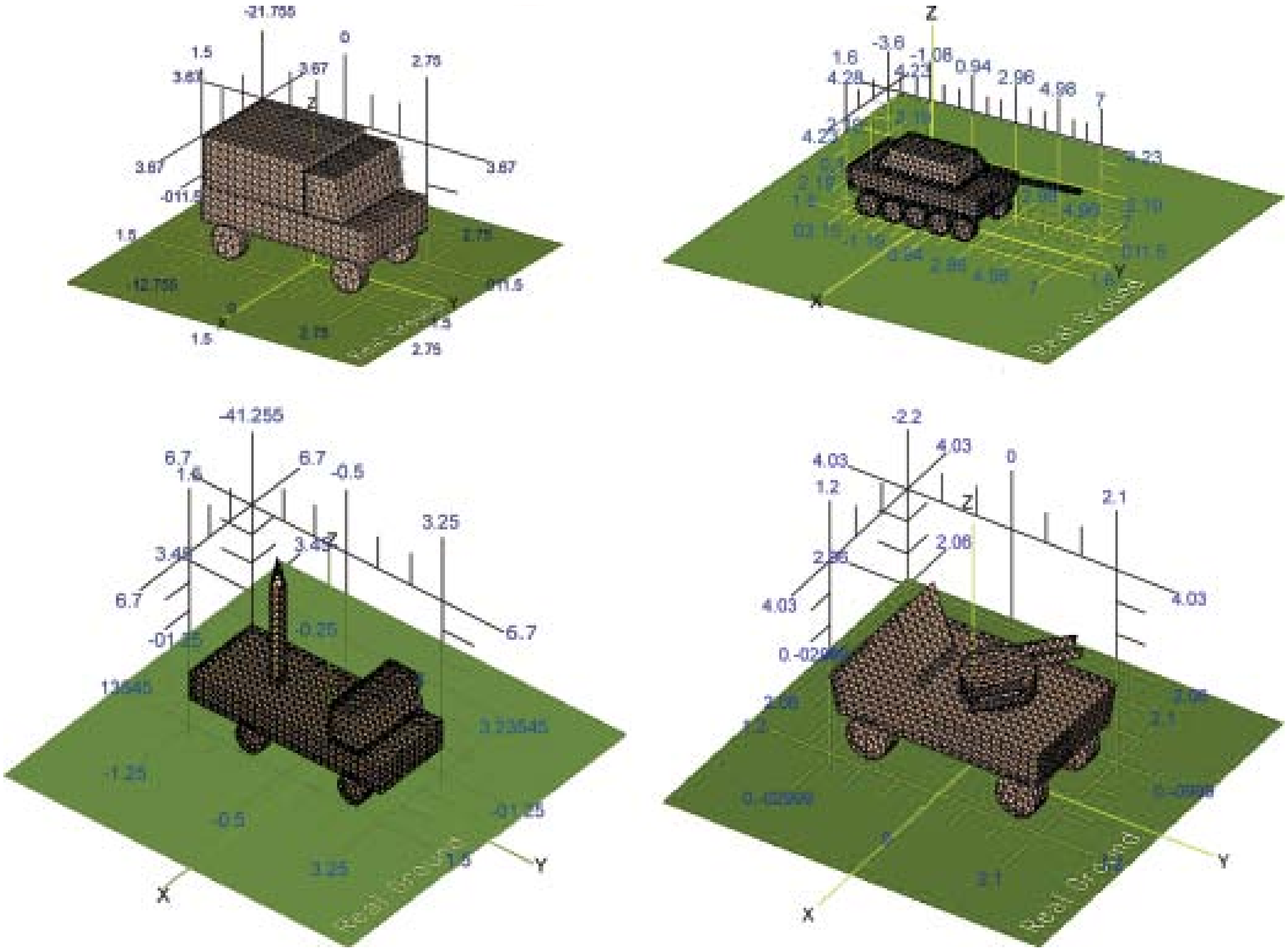

**Figure 5:** The CAD models of the four ground targets simulated in the present project (dimensions in meters).

or the guidance antenna of a land-to-air missile launcher.

Figure 5 shows the CAD models of the four ground targets, modeled in the current project. Dimensions of the targets have been kept close to that of real-life vehicles. As can be observed, the targets represent four of the major classes of vehicles which could commonly be found in ground combat environments.

Each target has been associated with the main features of the class, which it represents.

1) The *classifiable features* modeled for the armored personal carrier (APC) are the typical body shape, and a communication antenna in the front.
2) The *classifiable features* modeled for the main battle tank (MBT) are the turret, group of wheels, canon, and the communication antenna.
3) The *classifiable features* modeled for the stinger launcher (STR) are the battery of four stinger missiles, phased-array type guidance antenna, turret and a communication antenna.
4) The *classifiable features* modeled for the land missile launcher (MSL) are the truck-type body, a communication antenna and the attached big missile.

## 4. SAR Image Formation Steps

In the current article we analyze the task of simulating bistatic SAR images. This is because of two major reasons. First of all, bistatic radar system is an area of intense current research. And secondly, the simulation time taken for a bistatic SAR image formation is less than that taken for a monostatic SAR image formation. Because in the place of a moving transmitter and receiver, the transmitter is kept fixed in the current work. The simulation is done with a fixed transmitter, and the receiver revolving round the scene-to-be-imaged at a fixed elevation angle and with the azimuth angle varying from 0° to 360°. The samples collected by FEKO can be shown to form points in the k space [24-25] domain of the scene being imaged. The projection of the k space data sampled by this combination is as given by the shaded area in Figure 6. To form an image of the scene, small sub-regions are to be taken from this sampled k space. This is done by using the samples collected for a range of consecutive receiver azimuth positions. This k space data has a polar support, and hence the next step is to reformat the support into a rectangular-one by proper extrapolation of the sampled data points. The algorithm followed in the present work is that of keystone resampling [26]. After the polar-to-rectangular resampling, an inverse Fourier transform of the data grid gives the bistatic SAR image of the scene.

1 These are the classfiable features, as evident from the optical image of the models.

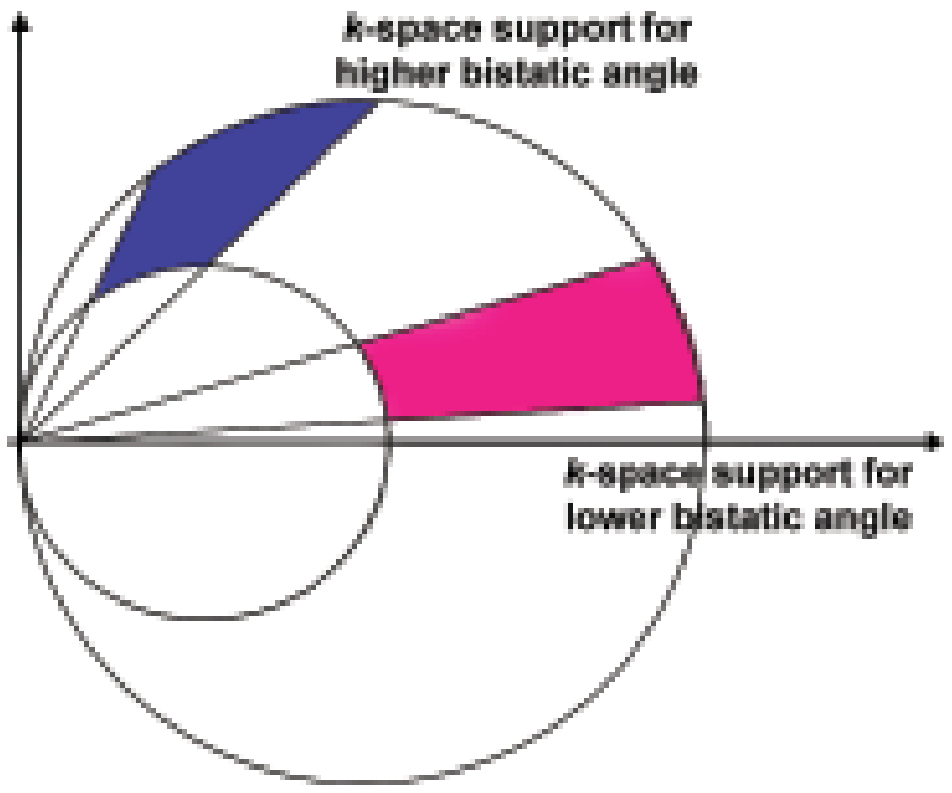

**Figure 6:** k space support is larger for lower bistatic angle.

A few points worth mentioning here are:

The resolution and hence the image quality of the generated bistatic SAR image, would depend on the k space support of the data, which in turn depends on the bistatic angle. Hence, the image resolution and the quality, depend on the bistatic angle of data collection.

For very high bistatic angles (as $\beta/2$ tends to 90°), as can be seen from Figure 6, the k space support nullifies, and the image quality is not acceptable.

The image formation steps can be summarized as follows:

The input parameters for the simulation are:

- Angle and frequency step;
- Bandwidth (BW) for which the EM simulator need to be run;
- CAD models of the targets.

With the above input parameters and files fed to the FEKO simulator, the transmitter was fixed at a certain azimuth and elevation, and the target was illuminated from the transmitter for the range of frequencies. This is done in frequency steps (as determined by the input parameters), covering the BW.

For a given transmitter position, and given frequency and polarization of transmission, the EM simulator is used to generate the surface current on the CAD model. This surface current file is stored.

Once the surface current file is known, the far field could be calculated for any given receiver azimuth and elevation. Hence the next step was to collect the scattered power from the target CAD model, for the range of frequencies, and with receiver position varying over a fixed elevation angle and a varying azimuth angle through 0° to 360°. This in turn is done at a few predetermined azimuth angular steps.

This gives the scattered field in both H and V polarizations of the receiver, for the range of frequencies and for the range of receiver azimuth angles. This forms an annular space in the k space as shown in the Figure 6 and is termed as data collected in *one run*.

The process was repeated for different positions of the transmitter and the receiver, and for different transmitter polarizations.

A total of four targets have been modeled in the current work.

For each transmitter position, scattered signals were collected for 360° of receiver azimuth round the target (one run) for two elevations: 10° and 15°. Hence, a total of 48 runs of data were collected for each polarization, for each target model.

Each run, as has been mentioned, gives the k space data for an annular ring as shown in Figure. From this, data are collected for a certain range of receiver angles for the BW of frequencies. This patch of k space data is polar-to-rectangular reformatted, as shown in the Figure 7 to make the k space support rectangular. Next the IFFT was taken to form the image of the target model. This image of the target model will display the target in the center of the image, and represents the SAR image of the target as obtained in real-SAR systems after discrimination phase [27]. This forms a target clip.[2]

2 A *target clip* is the SAR image of the target, with the image of the target at the center of the image. In an automatic target recognition exercise, after the detection stage, a particular part of the original SAR image of the scene is taken for further processing. This part of the SAR image, with the target at its center, is termed as the *target-clip*. In the present project, only the classification problem is handled. Hence, the input taken are the target image clips. Though no clipping operation is done, for convention, the word target clip is used through out the present report.

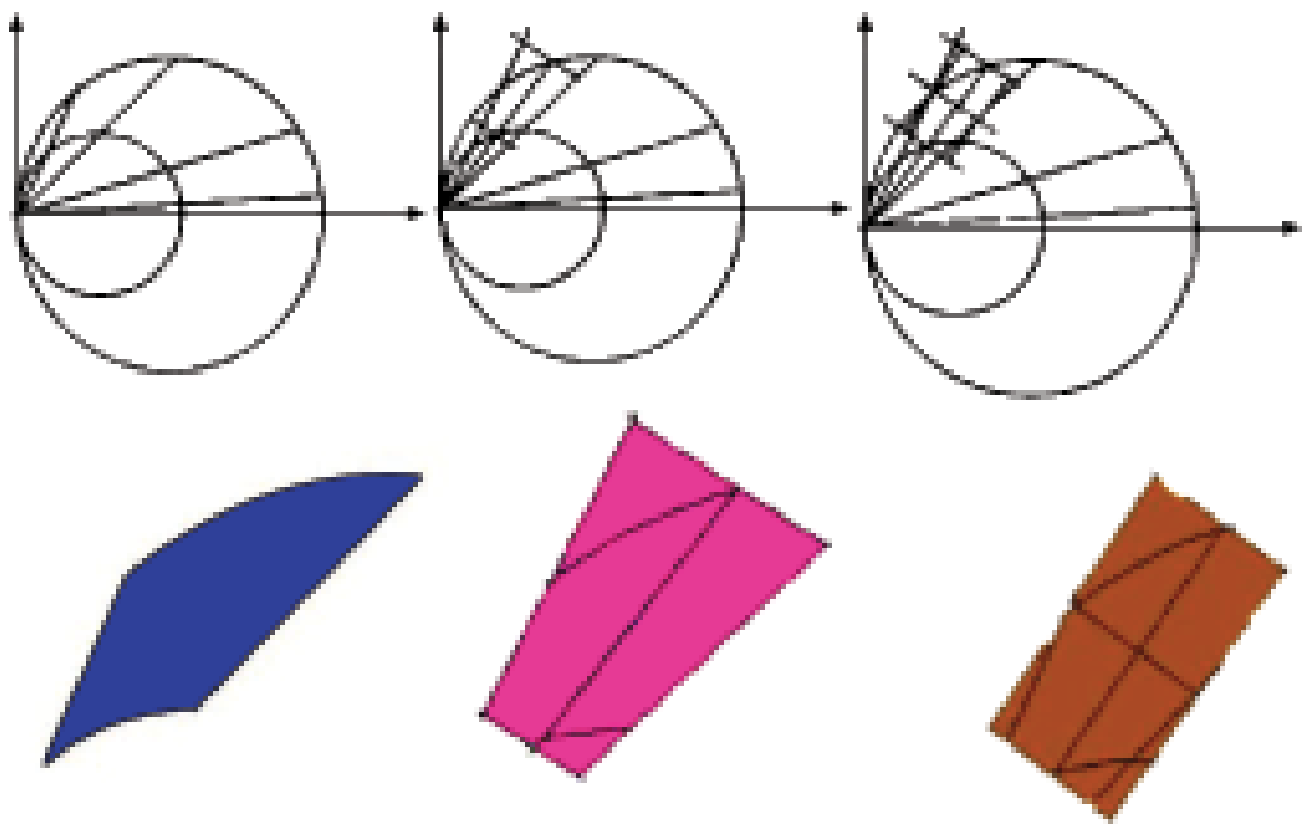

**Figure 7:** Key-stone approach of polar-to-rectangular reformatting of data collected.

With some overlapping of receiver azimuth angles for consecutive target clips, each run of data was used in the current project to generate around 25-50 image clips.

In a monostatic image formation exercise, the desired range resolution determines the bandwidth to be used, whereas the desired cross-range resolution determines the angular swath over which the range profiles are to be collected. However, in bistatic SAR imaging the image resolution depends on the bistatic angle of imaging [28-29]. To determine the bandwidth and the angular swath to be used for imaging in the current case, the best possible resolution using a bistatic case was taken as the landmark. Hence, most of the image clips will have a poorer resolution than the resolution figures used in finding the bandwidth of the system and the angular swath over which a single image clip is formed.

The central frequency was kept at 1 GHz. This decision was mainly controlled by the simulation constraints. In most of the EM-simulators, the model body is approximated as the summation of a triangular or rectangular mesh. The dimension of the triangles in the meshing must be an order smaller than the central frequency wavelength. Hence higher the central frequency, the smaller is the central frequency wavelength, and hence smaller the size of the triangles in the meshing. This results in a higher number of triangles required for the same model as compared to that required for a lower frequency. This heavily slows down the simulation speed. A center frequency of 1 GHz was found to be a compromise between our desired image resolution and a reasonable simulation speed. The spatial resolution was kept at around 0.2 m in both range and cross-range dimensions. For this requirement, the bandwidth was chosen to be 750 MHz and angular swath for each image formation was chosen to be 36°. All the target models were less than 8 m in dimension. Hence the maximum scene to be imaged was limited to around 10 m in both dimensions. This parameter was used to decide on the frequency step (15 MHz) and the angular step (0.72°).[3]

## 5. Analysis of a Sample Image

For a more detailed study of the SAR images generated in this project, this section analysis one of the images, in a more systematic way, to see how features in the image correspond to the physical features in the CAD model. Sometimes features are quite conspicuous, and sometimes difficult to identify. The aim of this analysis is twofold. First of all, it will be shown that all the features in the image have some link to the CAD model. This gives

3 The choice of bandwidth and central frequency makes the setup a wide-band simulation. Hence most of the derivations done earlier would not hold. However, in actual simulation, the complete bandwidth is discretised. For each simulation the bandwidth under consideration is just 15 MHz. Hence, the narrow band assumption still holds true for each simulation.

confidence in the images generated. Secondly, this also suggests that FEKO may be a competent tool for the task. For the current analysis, the bistatic SAR image of the model, *missile launcher* is chosen. The imaging parameters are, transmitter at 0° azimuth and 15° elevation, and receiver tracing an aperture of around 30° of azimuth starting from 0° to 30°, at a fixed elevation of 15°.

Figure 8 gives the optical and the bistatic SAR images of the target. The target in the scene, is a generic land to land missile launcher. The dimensions of the base of the target are 6.5 m in length, 3 m in width and 1m in height. The missile is 4 m in length and 0.5 m in diameter. A number of artifacts can be observed in the image, which may not be easily be linked to the physical features. To analyze the image and the correspondence of the image domain features to the physical features, features of the target were added one by one and the corresponding changes in image domain were observed.

First of all, the simplest feature, a perfectly electrical conducting (PEC) plate, was imaged Figure 9. As can be observed from the figure, the PEC plate behaves like four scattering centers, physically near the corners of the plate. Even though the exact position of these scattering centers could not be calculated, the SAR image in the figure is fairly accurate. Because of very low elevation angle, the scattering centers give very low return (as can be observed from the low amplitude spots in the image).

Next, the complete body and the head of the target were added [Figure 10]. Now, the return from the side plate of the body, which is perpendicular to the incident wave, is much stronger than the return from any other part.

This accounts for the brightest patch in the image.

In Figure 11, the wheels are added, and the changes in the SAR-image domain can be observed. Since the elevation angle is very low, the returns from all the wheels are visible to the receiver and hence form four bright spots in the image. Two of these spots lie close to the side plate and hence are not so conspicuous. The other two wheels to the rear, show up as significant bright spots.

Next, in Figure 12, the antenna element was added to one corner of the cabin (approximate position (1 m range and 2 m cross range)). A bright spot appears in almost the same position in the SAR image.

In Figure 13, the missile was added to the trolley of the vehicle. The corresponding appearance of a feature (a bright blob just above the left corner of the main bright patch) in SAR image can be observed clearly. As it has been explained before, this main bright patch represents the front edge of the target. It may be surprising to note that in SAR image domain, the dimension of the bright spot due to the missile is much smaller in size than what it looks in the optical image. In the model, the size of the missile is almost as long as the length of the vehicle. However, in the SAR image it appears just as a small blob. This is because SAR images are sometimes quite different from their optical counterparts. SAR images are inherently two dimensional and show the object as projected on the plane of incident EM waves. In the current case, the elevation angle of the illuminator platform was quite low (10°). The SAR image thus shows the object as projected on a plane tilted at an elevation angle of 10°. A rough ray diagram is shown in Figure 14. As can be seen from the ray diagram, the size of the missile after projection is much smaller than its original size.

Finally, comparing Figure 13 with Figure 8, the later has a ground plane added. This makes the return more strong and hence the overall image appears brighter and some faint artifacts are also observed in the image due to interaction with the ground plane.

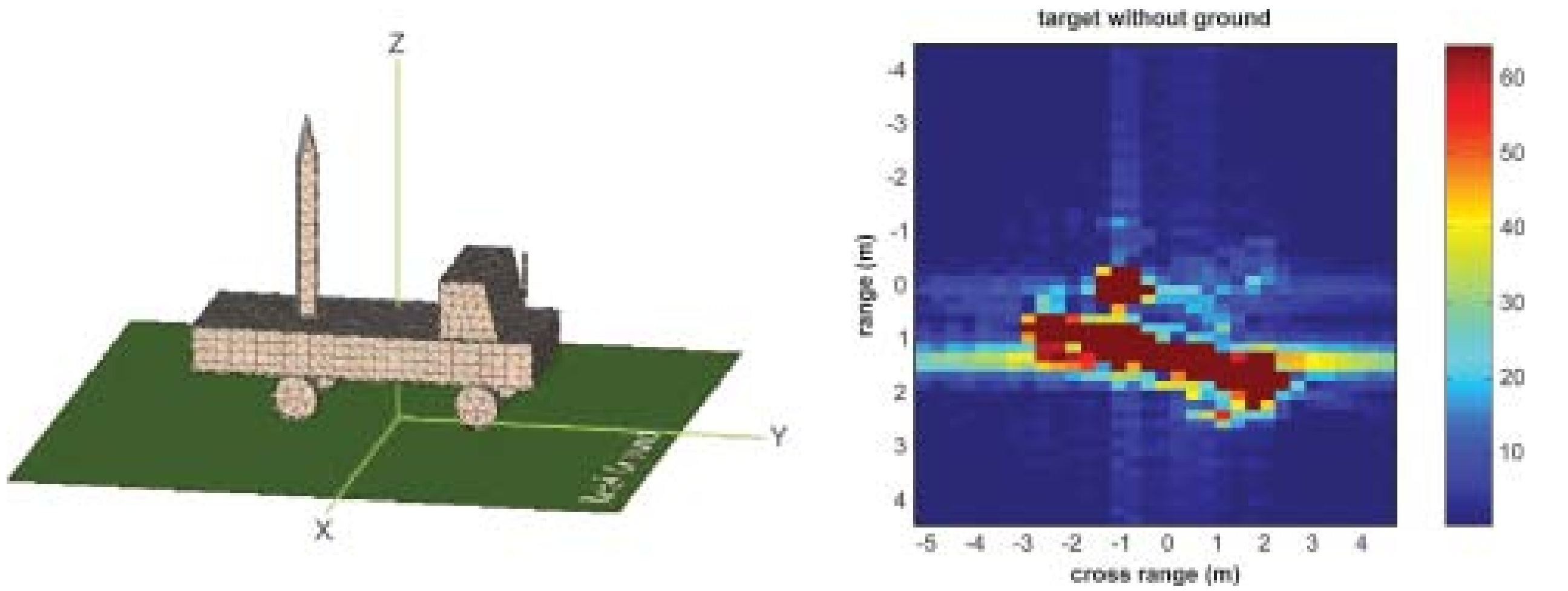

**Figure 8:** Optical and bistatic SAR images of the target on ground plane.

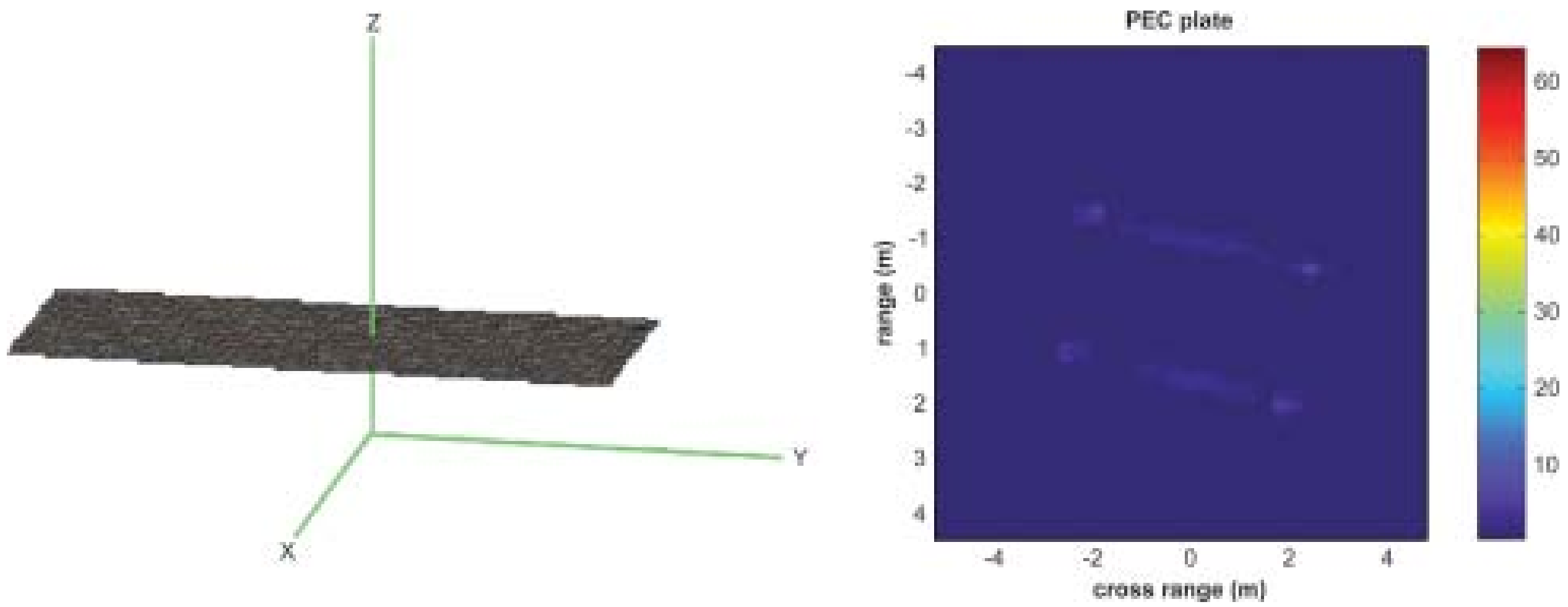

**Figure 9:** Optical and bistatic SAR images of the PEC flat plane.

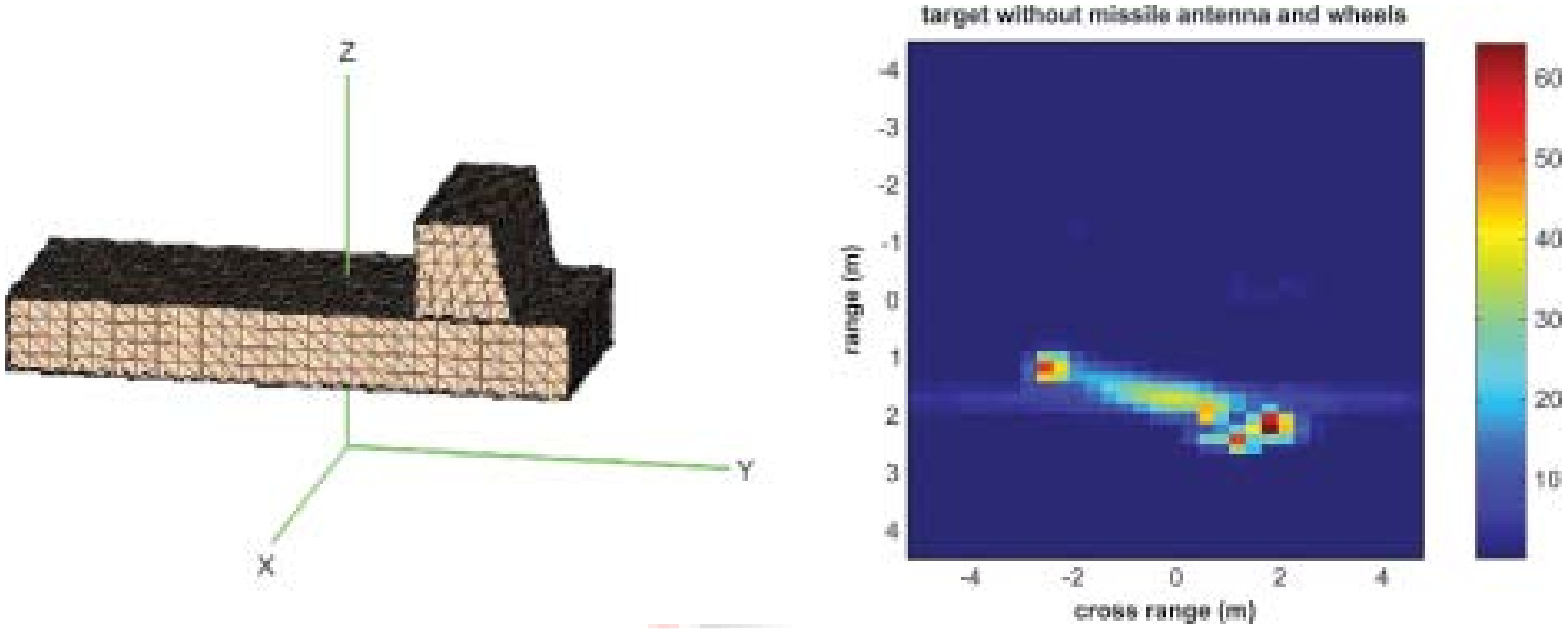

**Figure 10:** Optical and bistatic SAR images of the target with no wheels, no missile, no antenna and no ground plane.

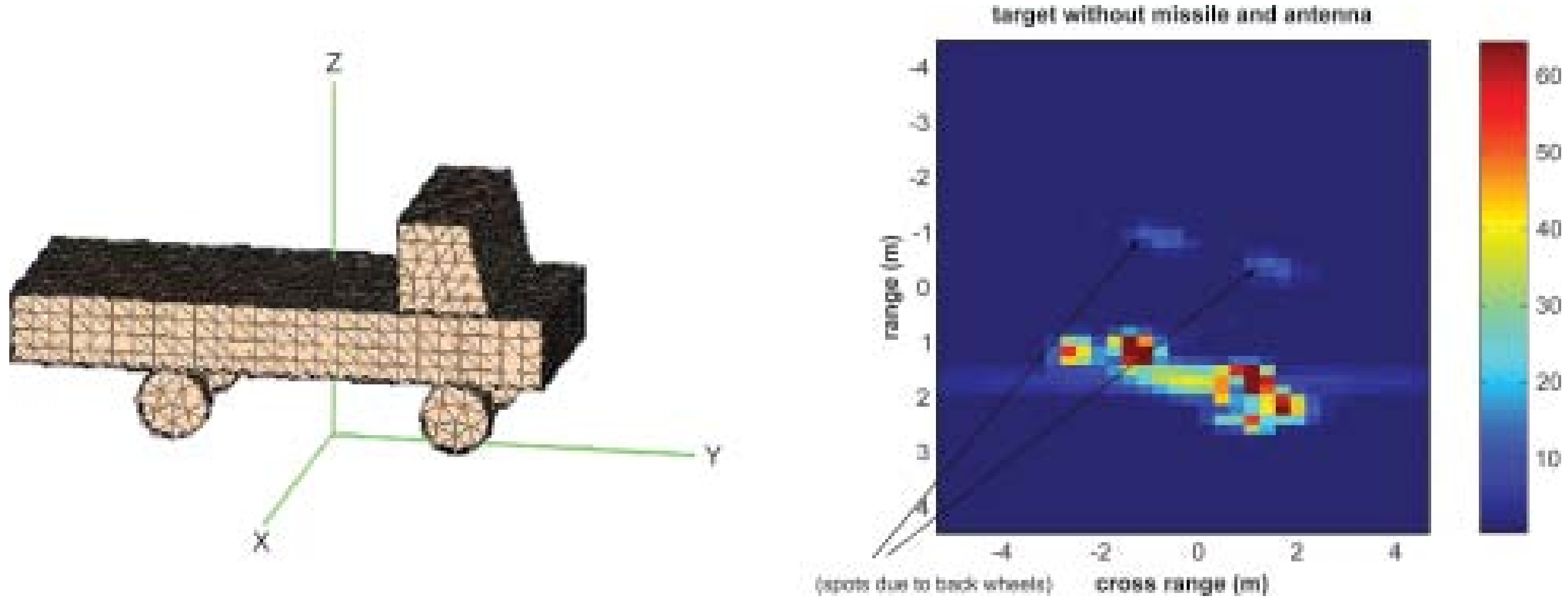

**Figure 11:** Optical and bistatic SAR images of the target with no missile, no antenna and no ground plane.

SAR imaging in a bistatic configuration is a complex procedure in itself, involving phenomena of different degrees of complexity. Using the simple physical optics (PO) method, features in the SAR image domain may not always appear so crisp and conspicuous when compared with the optical image. The above exercise was helpful in giving a fair insight into the SAR images and the correspondence of image domain features to the physical features. The typical bistatic SAR images of all the target models, (for one particular transmitter-receiver

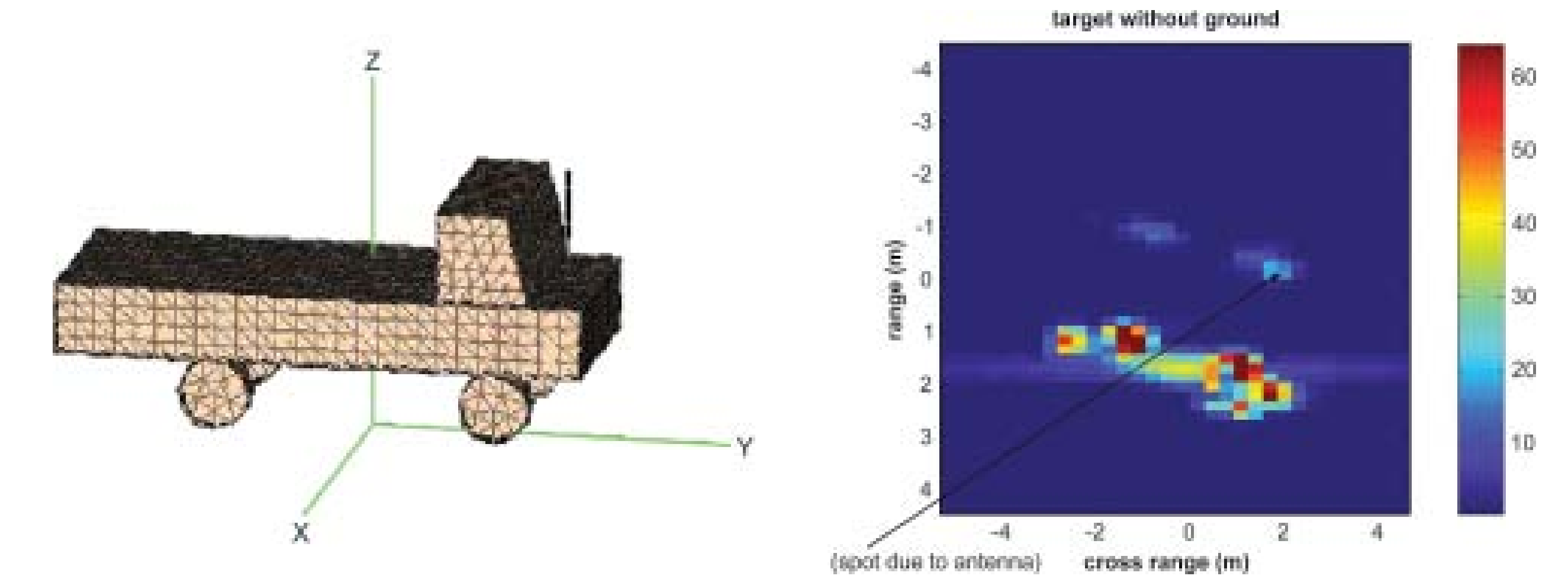

**Figure 12:** Optical and bistatic SAR images of the target with no missile, and no ground plane.

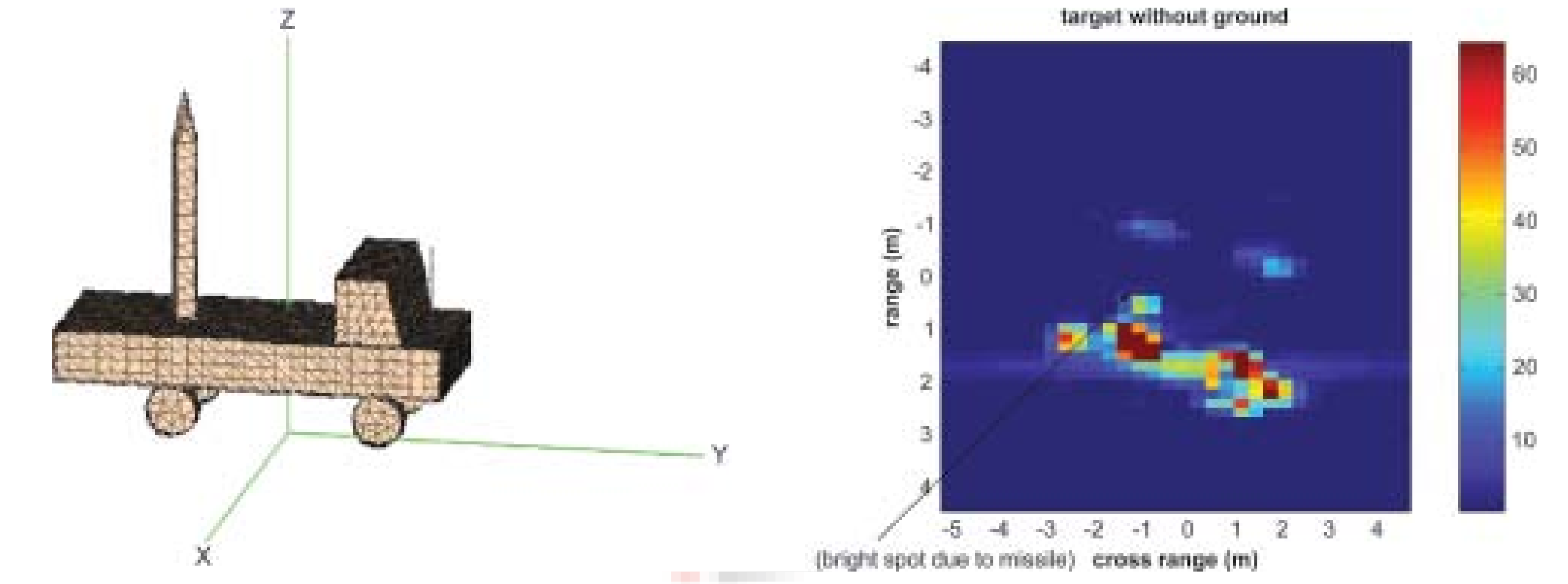

**Figure 13:** Optical and bistatic SAR images of the target with no ground plane.

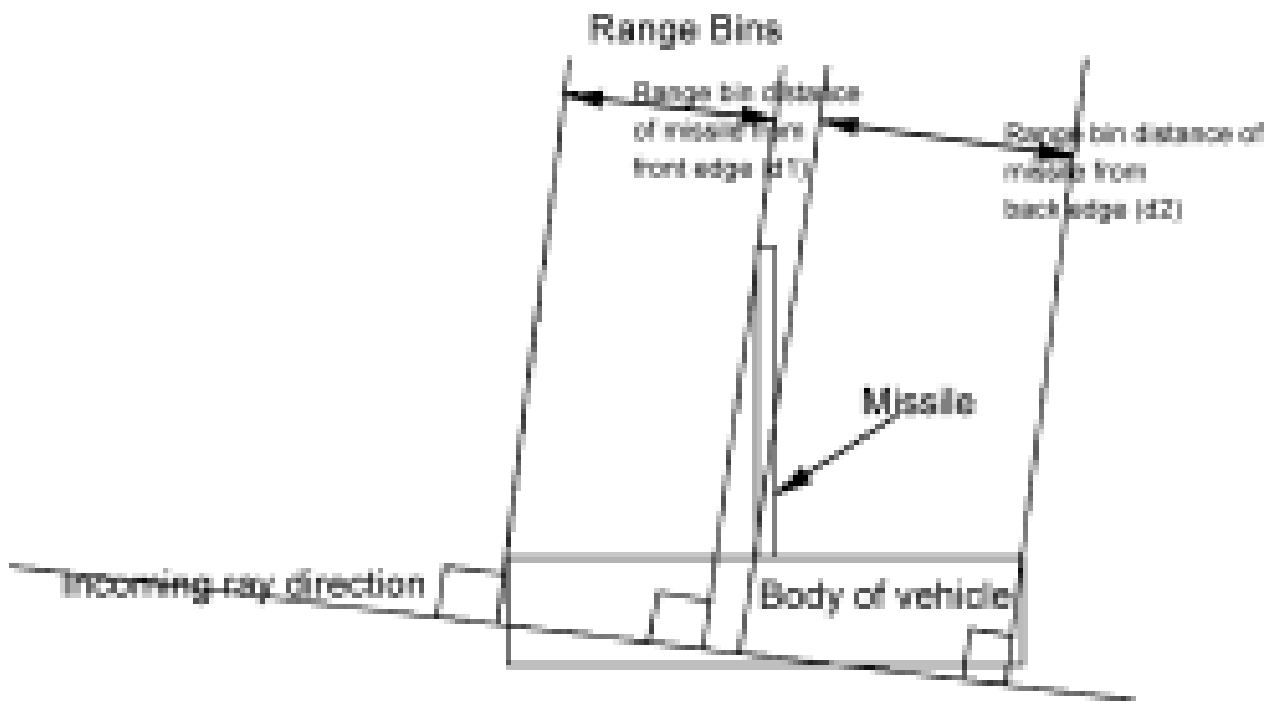

**Figure 14:** Ray diagram to show the reason for the proximity of the missile head to the front edge in SAR image.

combination) are shown in Figure 15.

## 6. Summary

The present work has been done almost from scratch, with almost no previous guidelines. Even though EM simulation packages like Epsilon and XPatch do give solutions to the problem of generating radar images of life sized vehicles, they are of restricted access. The present work is based on work done with the EM simulator FEKO. However it should be noted that FEKO is a general purpose EM modeling tool, mostly used for antenna design and EMC analysis. Secondly, no particular FEKO based feature has been used. All the features of FEKO used in the present work, can be found in any other general purpose EM simulator (for example WIPL-D [16]). Hence, the present work is almost tool independent, and can be of use to a much wider research group. One of the drawbacks of the present work is the lack of any standard validation. The novelty of the work put us in a situation where there were neither any standard results to compare our results against, nor were there any standard procedures to test such results. Hence, a lot of efforts have been put in understanding the results and validating them in whatever methods that were found suitable. The final results are of acceptable quality and have been used in one of the works by the authors on bistatic SAR ATR [30]. With the arrival of this new generation of modern EM simulation tools with immense power, it is expected that such simulation tools will be used more and more by different research groups in complimenting and supplementing their work on futuristic radar systems. With the increase in such efforts, the effort in modeling more detailed targets is also expected to grow. The author is of the opinion that soon we have to plan ahead to put forth standards in this field.

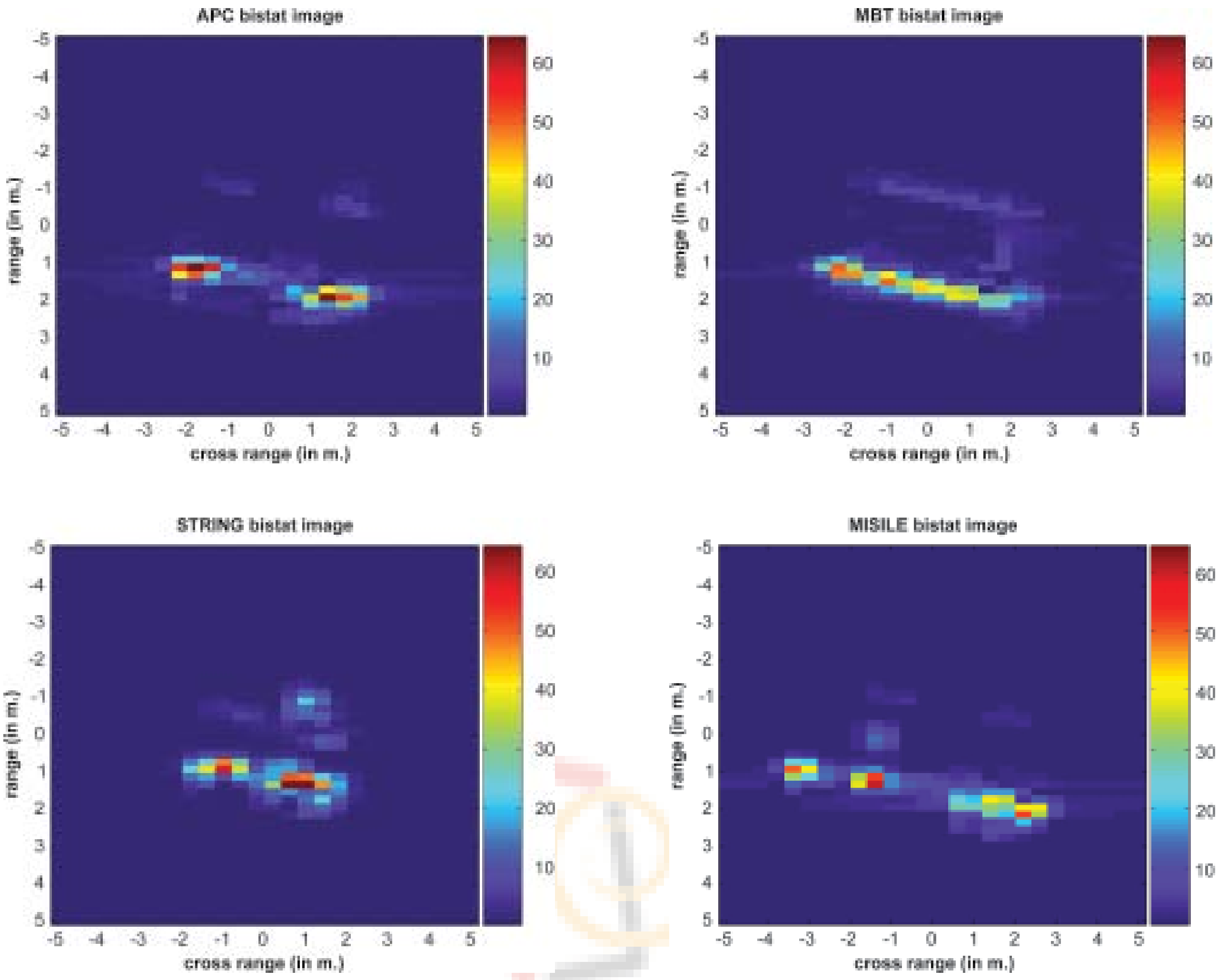

**Figure 15:** Representative bistatic SAR images of the four targets modeled (HH polarized images) (top left: APC, top right: MBT, bottom left: Missile launcher, bottom right: Stinger).

## 7. Acknowledgment and Disclaimer

The current report is based on the work done by the author in the university of Edinburgh, UK under a project from the electromagnetic remote sensing defense technology center (EMRSDTC). The author acknowledge the EMRSDTC for funding this project and the royal academy of engineers (RAE) for the equipment grant. Any view expressed are those of the author and do not necessarily represent those of the MOD or any other UK government department.

**Note: The color version of the figures are available in online version.**

## AUTHORS

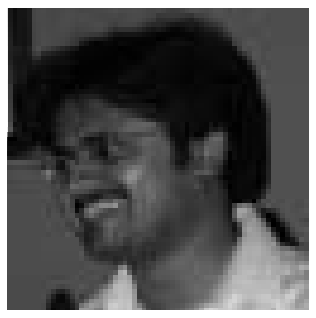

**Dr. Amit Kumar Mishra** received the B.E. Degree from National Institute of Technology Rourkela (formerly Regional Engineering College Rourkela), in 2001, and PhD, degree from University of Edinburgh, UK, in 2006. After receiving the B.E. Degree, he worked 1 year in defence R&D organization (DRDO), India and 1 year in Wipro Technologies. He has been with ECE department, Indian Institute of Technology Guwahati since 2006. His research interests are in pattern recognition, radar signal processing, VLSI DSP and ultrawideband (UWB) systems. He is the recipient of IETE (IRSI) young scientist award for the year 2008.

**E-mail:** akmishra@ieee.org

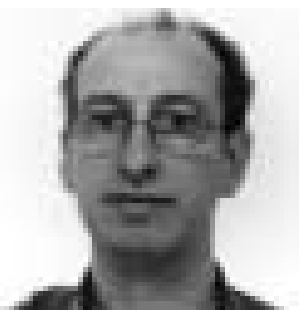

**Bernard (Bernie) Mulgrew** received the B.Sc. degree from Queen's University, Belfast, Ireland, in 1979, and Ph.D. degree from University of Edinburgh, Edinburgh, U.K., in 1986. After receiving the B.Sc. degree, he worked for 4 years as a Development Engineer in the Radar Systems Department at BAE Systems (formerly Ferranti), Edinburgh, U.K. From 1983 to 1986 he was a Research Associate in the Department of Electronics and Electrical Engineering at the University of Edinburgh, studying the performance and design of adaptive filter algorithms, and was appointed to a lectureship in 1987. He currently holds the BAE Systems/Royal Academy of Engineering Research Chair in Multi-Sensor Signal Processing and is Head of the Institute for Digital Communications at Edinburgh. His research interests are in adaptive signal processing and estimation theory and in their application to communications, radar and audio systems. He is a coauthor of three books on signal processing and over 50 journal papers. Dr. Mulgrew is a Fellow of the Institute of Electrical Engineers, U.K., a Fellow of the Royal Society of Edinburgh, and a member of the Audio Engineering Society.

**E-mail:** B.Mulgrew@ed.ac.uk